\documentclass[10pt,letterpaper,twocolumn]{article}
\usepackage[latin9]{inputenc}
\usepackage{color}
\usepackage{booktabs}
\usepackage{amsmath}
\usepackage{amssymb}
\usepackage{graphicx}

\makeatletter

\pdfpageheight\paperheight
\pdfpagewidth\paperwidth

\providecommand{\tabularnewline}{\\}

\usepackage{iccv}
\usepackage{times}
\usepackage{epsfig}
\usepackage[pagebackref=true,breaklinks=true,colorlinks,bookmarks=false]{hyperref}
\usepackage{booktabs}
\usepackage[sort,nocompress]{cite}
\usepackage{authblk}

\iccvfinalcopy 

\makeatother

\begin{document}
\author[1]{Eduardo P\'erez-Pellitero} 
\author[1]{Mehdi S.M.~Sajjadi} 
\author[2]{Michael Hirsch} 
\author[1,2]{Bernhard Sch\"olkopf}
\affil[1]{Max Planck Institute for Intelligent Systems} 
\affil[2]{Amazon Research}
\title{Perceptual Video Super Resolution with Enhanced Temporal Consistency}
\maketitle
\begin{abstract}
With the advent of perceptual loss functions, new possibilities in
super-resolution have emerged, and we currently have models that successfully
generate near-photorealistic high-resolution images from their low-resolution
observations. Up to now, however, such approaches have been exclusively
limited to single image super-resolution. The application of perceptual
loss functions on video processing still entails several challenges,
mostly related to the lack of temporal consistency of the generated
images, i.e.,~flickering artifacts. In this work, we present a novel
adversarial recurrent network for video upscaling that is able to
produce realistic textures in a temporally consistent way. The proposed
architecture naturally leverages information from previous frames
due to its recurrent architecture, i.e.~the input to the generator
is composed of the low-resolution image and, additionally, the warped
output of the network at the previous step. Together with a video
discriminator, we also propose additional loss functions to further
reinforce temporal consistency in the generated sequences. The experimental
validation of our algorithm shows the effectiveness of our approach
which obtains images with high perceptual quality and improved temporal
consistency.
\end{abstract}

\section{Introduction}

Advances in convolutional neural networks have revolutionized computer
vision and the popular field of super-resolution (SR) has been no
exception to this rule, as in recent years numerous publications have
made great strides towards better reconstructions of high-resolution
pictures. A most promising new trend in SR has emerged as the application
of \emph{perceptual} loss functions rather than the previously ubiquitous
optimization of the mean squared error. This paradigm shift has enabled
the leap from images with blurred textures to near-photorealistic
results in terms of perceived image quality using deep neural networks.
Notwithstanding the recent success in single image SR, perceptual
losses have not yet been successfully utilized in the video super
resolution (VSR) domain, as perceptual losses typically introduce
artifacts that, while being undisturbing in the spatial domain, emerge
as spurious flickering artifacts in videos.

In this paper we propose a neural network model that is able to produce
sharp videos with fine details while improving its behavior in terms
of temporal consistency. The contributions of the paper are: (1) A
recurrent generative adversarial model with a video discriminator,
(2) a multi-image warping that improves image alignment between adjacent
frames, and (3) two novel loss terms that reinforce temporal coherency
for consecutive frames.

\section{Related work}

The task of SR can be split into the groups of single image SR and
multi-frame or video SR methods.

Single image SR is one of the most relevant inverse problems in the
field of generative image processing tasks \cite{peyman,survey14}.
Since the initial work by Dong et al. \cite{srcnn} which applied
small convolutional neural networks to the task of single image SR,
several better neural network architectures have been proposed that
have achieved a significantly higher PSNR across various datasets
\cite{Shi2016,drcn,fastdong,deepresnetssr,lapsr,Lim2017,udnet}. Generally,
advances in network architectures for image detection tasks have also
helped in SR, e.g. adding residual connections \cite{residualnets}
enables the use of much deeper networks and speeds up training \cite{vdsr}.
We refer the reader to Agustsson and Timofte \cite{ntire} for a survey
of the state of the art in single image SR.

Since maximizing for PSNR leads to generally blurry images \cite{SajSchHir17},
another line of research has investigated alternative loss functions.
Johnson et al. \cite{johnson} and Alexey and Brox \cite{brox} replace
the mean squared error (MSE) in the image space with an MSE measurement
in feature space of large pre-trained image recognition networks.
Ledig et al. \cite{Ledig2017} extend this idea by adding an adversarial
loss and Sajjadi et al. \cite{SajSchHir17} combine perceptual, adversarial
and texture synthesis loss terms to produce sharper images with hallucinated
details. Although these methods produce detailed images, they typically
contain small artifacts that are visible upon close inspection. While
such artifacts are bearable in images, they lead to flickering in
super-resolved videos. For this reason, applying these perceptual
loss functions to the problem of video SR is more involved.

Amongst classical video SR methods, Liu et al. \cite{bayesianadaptive}
have achieved notable image quality using Bayesian optimization methods,
but the computational complexity of the approach prohibits use in
real-time applications. Neural network based approaches include Huang
et al. \cite{bidirectional} who use a bidirectional recurrent architecture
with comparably shallow networks without explicit motion compensation.
More recently, neural network based methods operate on a sliding window
of input frames. The main idea of Kappeler et al.~\cite{vsrnet}
is to align and warp neighboring frames to the current frame before
all images are fed into a SR network which combines details from all
frames into a single image. Inspired by this idea, Caballero et al.~\cite{Caballero2017}
take a similar approach but employ a flow estimation network for the
frame alignment. Similarly, Makansi et al.~\cite{endtoendmotion}
use a sliding window approach but they combine the frame alignment
and SR steps. Tao et al. \cite{detailrevealing} also propose a method
which operates on a stack of video frames. They estimate the motion
in the frames and subsequently map them into high-resolution space
before another SR network combines the information from all frames.
Liu et al. \cite{robustdyn} operate on varying numbers of frames
at the same time to generate different high-resolution images and
then condense the results into a single image in a final step.

For generative video processing methods, temporal consistency of the
output is crucial. Since most recent methods operate on a sliding
window \cite{Caballero2017,detailrevealing,robustdyn,endtoendmotion},
it is hard to optimize the networks to produce temporally consistent
results as no information of the previously super-resolved frame is
directly included in the next step. To accommodate for this, Sajjadi
et al. \cite{Sajjadi2018} use a frame-recurrent approach where the
estimated high-resolution frame of the previous step is fed into the
network for the following step. This encourages more temporally consistent
results, however the authors do not explicitly employ a loss term
for the temporal consistency of the output.

To the best of our knowledge, VSR methods have so far been restricted
to MSE optimization methods and recent advancements in perceptual
image quality in single image SR have not yet been successfully transferred
to VSR. A possible explanation is that perceptual losses lead to sharper
images which makes temporal inconsistencies significantly more evident
in the results, leading to unpleasing flickering in the high-resolution
videos \cite{SajSchHir17}.

The style transfer community has faced similar problems in their transition
from single-image to video processing. Single-image style-transfer
networks might produce very distant images for adjacent frames \cite{Gatys2016},
creating very strong transitions from frame to frame. Several recent
works have overcome this problem by including a temporal-consistency
loss that ensures that the stylized consecutive frames are similar
to each other when warped with the optical flow of the scene \cite{Gupta2017,Ruder2016,Huang2017}.

In this work, inspired by the contributions above, we explore the
application of perceptual losses for VSR using adversarial training
and temporal consistency objectives. 

\section{Proposed method\label{sec:Proposed-method}}

\begin{figure*}
\begin{centering}
\includegraphics[width=1\textwidth]{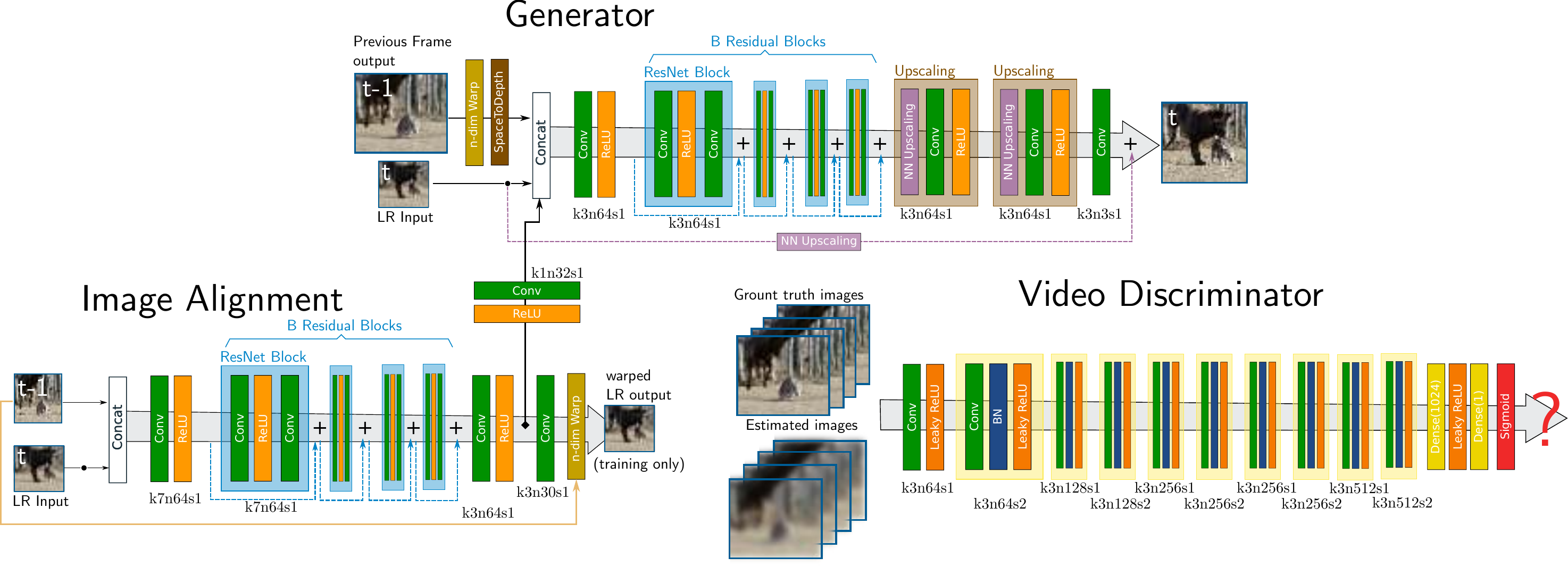}
\par\end{centering}
\caption{Network architectures for generator and discriminator. The previous
output frame is warped onto the current frame and mapped to LR with
the space to depth transformation before being concatenated to the
current LR input frame. The generator follows a ResNet architecture
with skip connections around the residual blocks and around the whole
network. The discriminator follows the common design pattern of decreasing
the spatial dimension of the images while increasing the number of
channels after each block. \label{fig:RG_and D}}
\end{figure*}

\subsection{Notation and problem statement}

VSR aims at upscaling a given LR image sequence $\left\{ Y_{t}\right\} $
by a factor of $s$, so that the estimated sequence \big\{$\tilde{X}_{t}$\big\} 
resembles the original sequence $\left\{ X_{t}\right\} $ by some
metric. We denote images in the low-resolution domain by $Y\in[0,1]^{h\times w\times3}$,
and ground-truth images in the high-resolution domain by $X\in[0,1]^{sh\times sw\times3}$
for a given magnification factor $s$. An estimate of a high-resolution
image $X$ is denoted by $\tilde{X}$. We discern within a temporal
sequence by a subindex to the image variable, e.g., $Y_{t-1},\:Y_{t}$.
We use a superscript $w$, e.g.~$\tilde{X}_{t-1}^{w}$, to denote
an image $\tilde{X}$ that has been warped from its time step $t-1$
to the following frame $X_{t}$.

The proposed architecture is summarized in Figure~\ref{fig:RG_and D}
and will be explained in detail in the following sections. We define
an architecture that naturally leverages not only single image but
also inter-frame details present in video sequences by using a recurrent
neural network architecture. The previous output frame is aligned
through a image alignment network. By including a video discriminator
that is only needed at the training stage, we further enable adversarial
training which has proved to be a powerful tool for generating sharper
and more realistic images \cite{SajSchHir17,Ledig2017}.

To the best of our knowledge, the use of perceptual loss functions
(i.e.~adversarial training in recurrent architectures) for VSR is
novel. 

\subsection{Recurrent generator and video discriminator\label{subsec:Recurrent-generator-and}}

Following recent SR state of the art methods for both classical and
perceptual loss functions \cite{Kim2016,Lim2017,SajSchHir17,Ledig2017},
we use deep convolutional neural networks with residual connections.
This class of networks facilitates learning the identity mapping and
leads to better gradient flow through deep networks. Specifically,
we adopt a ResNet architecture for our recurrent generator that is
similar to the ones introduced by \cite{Ledig2017,SajSchHir17} with
some modifications.

Each of the residual blocks is composed by a convolution, a Rectified
Linear Unit (ReLU) activation and another convolutional layer following
the activation. Previous approaches have applied batch normalization
layers in the residual blocks \cite{Ledig2017}, but we choose not
to add batch normalization to the generator due to the comparably
small batch size, to avoid potential color shift problems, and also
taking into account recent evidence hinting that they might be problematic
for generative image models \cite{Xiang2017}. In order to further
accelerate and stabilize training, we create an additional skip connection
over the whole generator. This means that the network only needs to
learn the residual between the nearest neighbor interpolation of the
input and the high-resolution ground-truth image rather than having
to pass through all low frequencies as well \cite{SajSchHir17,deepresnetssr}.

We perform most of our convolutions in low-resolution space for a
higher receptive field and higher efficiency. Since the input image
has a lower dimension than the output image, the generator needs to
have a module that increases the resolution towards the end. There
are several ways to do so within a neural network, e.g., transposed
convolution layers, interpolation or depth to space units \textit{(pixelshuffle)}.
In order to avoid potential grid artifacts when introducing the adversarial
loss, we decided to perform the upscaling via nearest neighbor interpolation.
The upscaling unit is divided into two stages with an intermediate
magnification step $r$ (e.g.\ two times $\times2$ for a magnification
factor of $\times4$). Each of the upscaling stages is composed of
a nearest neighbor interpolation, a convolutional layer and a ReLU
activation.

In contrast to general generative adversarial networks, the input
to the proposed generative network is not a random variable but it
is composed of the low-resolution image $Y_{t}$ (corresponding to
the current frame $t$) and, additionally, the warped output of the
network at the previous step $\tilde{X}_{t-1}^{w}$. The difference
in resolution of these two images is adapted through a space to channel
layer which decreases the spatial resolution of $\tilde{X_{t-1}^{w}}$
without loss of information. As for the previous-image warping approach,
we propose using an $n$-dimensional optical flow field that is estimated
with a separate, non-recurrent network (refer to Section~\ref{subsec:Resampling-and-alignment}). 

Our discriminator follows common design choices and is composed of
strided convolutions, batch normalization and leaky ReLU activations
that progressively decrease the spatial resolution of the activations
while increasing the channel count \cite{Ledig2017,Radford2016,SajSchHir17}.
The last stage of the discriminator is composed of two dense layers
and a sigmoid activation function. Differently to single-image SR
approaches, we let the discriminator see the stream of images produced
by the unfolded generator, e.g.~10 images. This enables the discriminator
to also evaluate temporal consistency in the classification of fake
and real sequences.

\subsection{Resampling and alignment between frames\label{subsec:Resampling-and-alignment}}

Many classic vision tasks do not deal with single images but rather
with streams of images that expand the temporal dimension. Optical
flow has been widely used as a representation that describes temporal
relationships within images, and thus enables temporal cues in learning.
Optical flow represents the motion perceived by the camera by two
image fields $(u,v)$, where each element describes the pixel-wise
vertical and horizontal displacement. Recently, there have been several
contributions on optical flow estimation via deep neural networks,
e.g. FlowNet \cite{Dosovitskiy:2015:FLO:2919332.2919957}, SPyNet
\cite{spynet}. The later method use the resampling operation (i.e.~bilinear
warping) in order to progressively refine its flow estimates. Within
VSR, most of the motion-aware methods \cite{Sajjadi2018,Li2017} use
optical flow fields to warp and align images. In VSR, however, motion
compensation networks are generally trained on warping image error,
and the decisive factor is not so much how accurate the flow fields
are but rather how accurate the warped image is. Also, often the warped
and aligned image is used for inference and the information of the
flow fields is otherwise discarded, even though it might carry meaningful
motion information.

\begin{figure}
\begin{centering}
\includegraphics[scale=0.5]{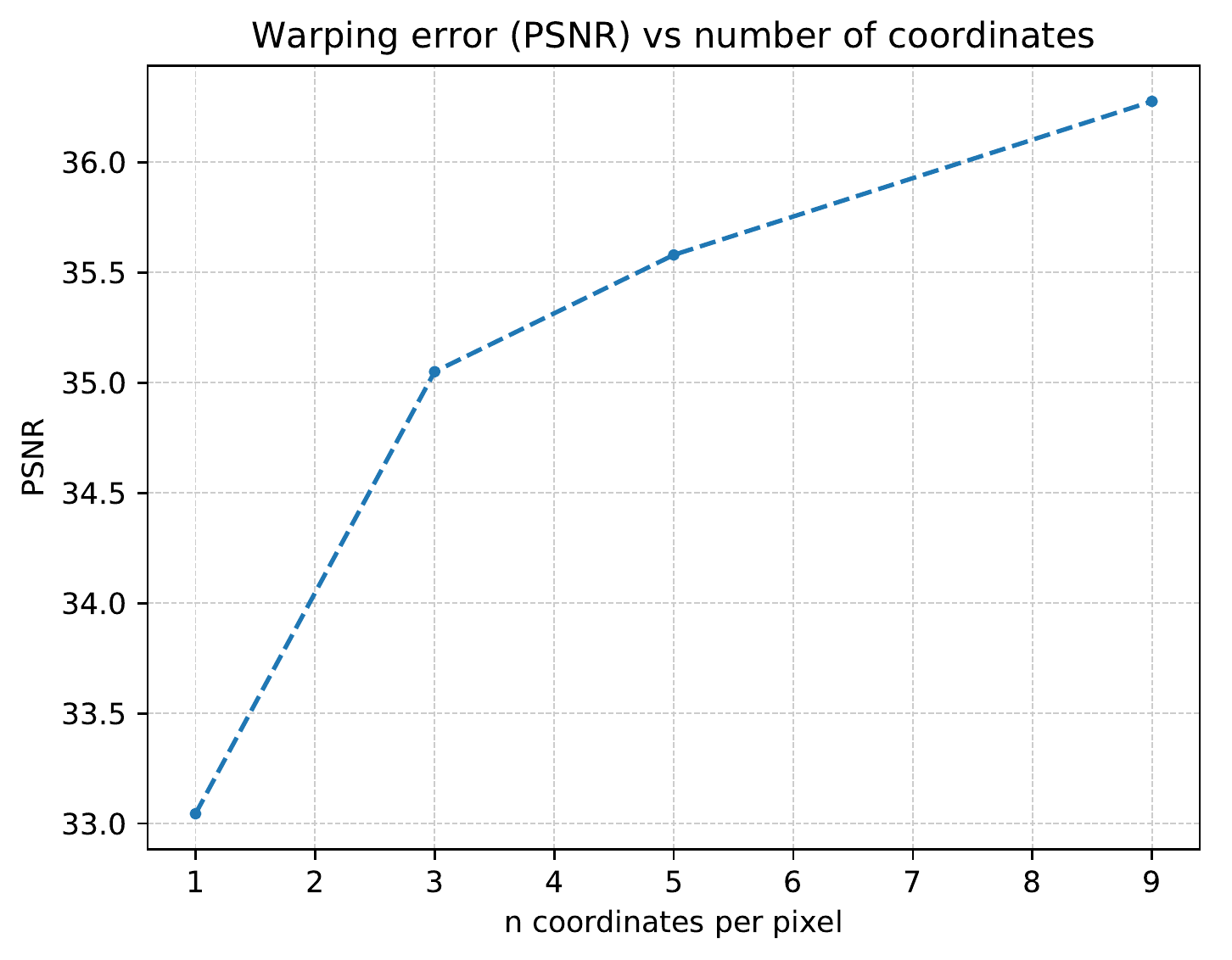}
\par\end{centering}
\caption{Warping error PSNR vs number of coordinates used for the image alignment
(see Equation \ref{eq:n_warping}) for the seq12 dataset. Previous
works only estimate and apply $n=1$, which corresponds to traditional
optical flow $(u,v)$. By increasing the number of estimated coordinates
and linearly combining their correspondent pixel values the warping
error is greatly reduced.\label{fig:ablation_ndim}}

\end{figure}

In this paper we propose a new paradigm for warping and aligning images.
Traditional optical flow algorithms estimate just one coordinate per
input pixel. In this paper, we propose estimating $n$ coordinates
whose correspondent pixel values are then linearly combined via a
set of corresponding $n$ weights. The warped image $\tilde{X}_{w}^{t-1}$
is obtained as follows:

\begin{equation}
\tilde{X}_{w}^{t-1}=\sum_{i=1}^{n}w_{i}\text{\ensuremath{\odot}}\tilde{X}^{t-1}(x+u_{i},y+v_{i}),\label{eq:n_warping}
\end{equation}

where operator $\odot$ performs element-wise multiplication, $X(x,y)$
samples the image $X$ at spatial locations $(x,y)$ and $w_{i}$
is a matrix of weights matching the image size. The proposed multi-image
warping scheme allows each warped pixel to be composed by a non-rigid
linear combination of several pixels, thus being more expressive and
also effectively bypassing the limiting factor of the bilinear interpolation
as the last step of the image warping, i.e.~the network can improve
the final image by combining the $n$ intermediate images. We show
warping accuracy with respect to $n$ in Figure~\ref{fig:ablation_ndim}.

In our architecture (see Figure~\ref{fig:RG_and D}) we also bridge
the last feature activation of our motion compensation network to
the image generator in order to provide motion-aware features to the
generator and to improve information flow (i.e.~as opposed to only
passing the warped image). 

\subsection{Losses\label{subsec:Losses}}

Upscaling video sequences has the additional challenge of respecting
the original temporal consistency between adjacent frames so that
the estimated video does not present unpleasing flickering artifacts.

When minimizing only MSE such artifacts are less noticeable for two
main reasons: because (1) MSE minimization often converges to the
mean in textured regions, and thus flickering is reduced and (2) the
pixel-wise MSE with respect to the ground truth (GT) is up to a certain
point enforcing the inter-frame consistency present in the training
images. However, when adding an adversarial loss term, the difficult
to maintain temporal consistency increases. Adversarial training aims
at generating samples that lie in the manifold of images, and thus
it generates high-frequency content that will hardly be pixel-wise
accurate to any ground-truth image. 

The architecture presented in Section~\ref{subsec:Recurrent-generator-and}
is naturally able to learn temporal dependencies thanks to its recurrent
design and its multi-frame discriminator. We train it with L1, texture
and adversarial losses. Additionally, we introduce two novel losses
in order to further reinforce temporal consistency. 

\subsubsection{L1 distance}

MSE is by far the most common loss in the SR literature as it is well-understood
and easy to compute. It accurately captures sharp edges and contours,
but it leads to over-smooth and flat textures as the reconstruction
of high-frequency areas falls to the local mean rather than a realistic
mode \cite{SajSchHir17}. Recently, also L1 distance (absolute error)
has been used for image restoration, as it behaves similarly to the
well-known L2 distance and it has been reported to perform slightly
better than L2 distance.

The pixel-wise L1 distance is defined as follows:

\begin{equation}
\mathcal{L}_{E}=\left\Vert \tilde{X}_{t}-X_{t}\right\Vert _{1},\label{eq:L_e}
\end{equation}

where $\tilde{X}_{t}$ denotes the estimated image of the generator
for frame $t$ and $X_{t}$ denotes the ground-truth HR frame $t$.

\subsubsection{Adversarial Loss}

\begin{figure}[t]
\begin{centering}
\includegraphics[width=1\columnwidth]{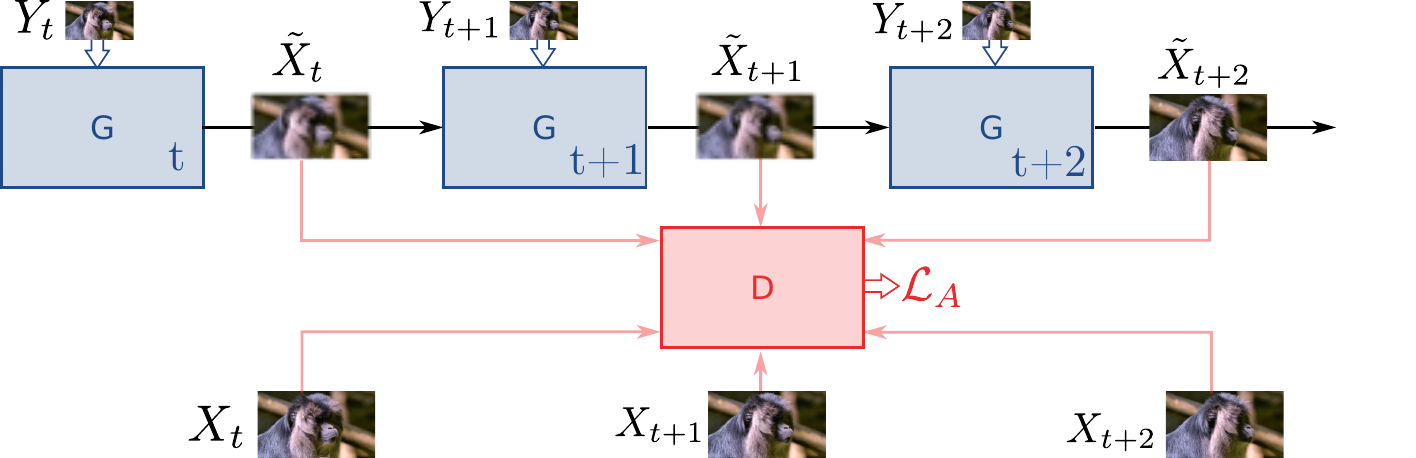}
\par\end{centering}
\caption{Unfolded recurrent generator $G$ and discriminator $D$ during training
for 3 temporal steps. The output of the previous time step is fed
into the generator for the next iteration. Note that the weights of
$G$ are shared across different time steps. Gradients of all losses
during training pass through the whole unrolled configuration of network
instances. The discriminator receives as input as many images as temporal
steps are during training. \label{fig:Unrolled-recurrent-network}}
\end{figure}
Generative Adversarial Networks (GANs)~\cite{Goodfellow2014} and
their characteristic adversarial training scheme have been a very
active research field in the recent years, defining a wide landscape
of applications. In GANs, a generative model is obtained by simultaneously
training an additional network. A generative model $G$ (i.e.~generator)
that learns to produce samples close to the data distribution of the
training set is trained along with a discriminative model $D$ (i.e.~discriminator)
that estimates the probability of a given sample belonging to the
training set or not, i.e., it is generated by $G$. The objective
of $G$ is to maximize the errors committed by $D$, whereas the training
of $D$ should minimize its own errors, leading to a two-player minimax
game.

Similar to previous single-image SR \cite{Ledig2017,SajSchHir17},
the input to the generator $G$ is not a random vector but an LR image
(in our case, with an additional recurrent input), and thus the generator
minimizes the following loss:

\begin{equation}
\mathcal{L}_{A}=-\log(D(G(Y_{t}\vert\vert\tilde{X}_{t-1})),
\end{equation}
where the operator $\vert\vert$ denotes concatenation. The discriminator
minimizes:

\vspace{-2.5ex}
\begin{equation}
\mathcal{L}_{D}=-\log(D(X_{t}))-\log(1-D(G(Y_{t}\vert\vert\tilde{X}_{t-1})).
\end{equation}

In this specific discriminator set-up, we would like to remark that
the adversarial loss enforces temporal consistency as well, as differently
to other single-image SR, in our architecture the discriminator has
access to multiple frames and thus can leverage information contained
over the temporal dimension in order to classify its inputs.

\subsubsection{Texture Loss}

For the purpose of image style transfer, Gatys et al.~\cite{Gatys2016}
found that feature correlations from pre-trained convolutional neural
networks capture the \textit{style} of images which can then be used
for realistic texture synthesis. Sajjadi et al.~\cite{Sajjadi2018}
propose to apply this method for SR in oder to match the texture of
the generated images to the original high-resolution textures at training
time. To this end, we compute VGG \cite{Simonyan15} features $\phi$
for both ground-truth images${X_{t}}$ and generated images ${\tilde{X}_{t}}$
and compute the corresponding gram matrices $G(F)=FF^{T}$. The final
loss term reads:

\begin{equation}
\mathcal{L}_{G}=||G(\phi(\tilde{X_{t}}))-G(\phi(X_{t}))||_{1}
\end{equation}

The texture loss term encourages sharper and more accurate textures
in the generated videos and furthermore stabilizes training, which
is important given the recurrence loop. 

\subsubsection{Static Temporal Loss}

When warping an image to compensate its motion, most of its high-frequency
content is filtered out (as the warping operation behaves as a low-pass
filter in the frequency domain) and thus, comparing warped images
is not effective in evaluating or avoiding flickering artifacts. Additionally,
flickering artifacts are most noticeable when they occur in regions
of the video that are still (i.e.~there is not motion across frames).
For that purpose, we propose the static temporal loss $\mathcal{L}_{Td}$.
This loss computes the difference across frames (without warping)
only for regions where there is no pixel-value variation in the ground-truth
images. First, we compute the following mask with the ground-truth
images:

\begin{equation}
m_{t}=\exp(-\alpha\left\Vert X_{t}-X_{t-1}\right\Vert _{2}^{2}),
\end{equation}
where $\alpha$ is sufficiently large to have fast transitions from
$1$ to $0$ whenever the frame difference is non-zero (we fixed $\alpha=100)$.
We compute then the distance between consecutive estimated images
and apply $m_{t}$ to it, thus obtaining an loss signal for those
regions that should remain static:

\begin{equation}
\mathcal{L}_{T_{d}}=m_{t}\ensuremath{\odot}\left\Vert \tilde{X}_{t}-\tilde{X}_{t-1}\right\Vert _{1}.\label{eq:L_td}
\end{equation}

\subsubsection{Temporal Statistics Loss}

In order to reproduce the temporal characteristics of the original
sequence without any direct pixel-wise comparison we compute the variance
over the temporal dimension and match the statistics of the estimated
images to those of the original sequence.

We compute at each image location $(x,y)$ the variance across time
both for GT images $\sigma^{2}(x,y)={\textstyle var}(\left\{ X_{t}(x,y)\right\} )$
and for the estimated images $\tilde{\sigma}^{2}(x,y)={\textstyle var}(\left\{ \tilde{X}_{t}(x,y)\right\} )$.
Those statistics represent how much variation is there in a given
image location across time, and thus is representative of the temporal
consistency at each pixel location. We compute the loss term as follows:

\begin{equation}
\mathcal{L}_{T_{s}}=\left\Vert \sigma^{2}-\tilde{\sigma}^{2}\right\Vert _{1}.\label{eq:L_Ts}
\end{equation}

\section{Results}

\subsection{Training and parameters}

\begin{table*}
\begin{centering}
\begin{tabular}{lrrrrrrrrrrr}
\toprule 
 & \multicolumn{5}{c}{\textit{\footnotesize{}Pixel-error objective}} & \multicolumn{6}{c}{\textit{\footnotesize{}Perceptual objective}}\tabularnewline
\cmidrule{2-12} \cmidrule{3-12} \cmidrule{4-12} \cmidrule{5-12} \cmidrule{6-12} \cmidrule{7-12} \cmidrule{8-12} \cmidrule{9-12} \cmidrule{10-12} \cmidrule{11-12} \cmidrule{12-12} 
 & {\footnotesize{}bicubic} & {\scriptsize{}$B_{1,2,3}+T$} & {\footnotesize{}DRDVSR} & {\footnotesize{}FRVSR} & {\footnotesize{}$\mathcal{L}_{E}$} & {\footnotesize{}ENet} & {\footnotesize{}SRGAN} & {\footnotesize{}$\mathcal{L}_{A}$} & {\footnotesize{}$\mathcal{L}_{T_{d}}$} & {\footnotesize{}$\mathcal{L}_{T_{s}}$} & {\footnotesize{}$\mathcal{L}_{C}$}\tabularnewline
\midrule 
{\footnotesize{}PSNR} & {\footnotesize{}22.443} & {\footnotesize{}23.898} & {\footnotesize{}24.389} & \textcolor{blue}{\footnotesize{}25.210} & \textbf{\footnotesize{}25.226} & {\footnotesize{}20.886} & {\footnotesize{}19.783} & {\footnotesize{}22.079} & \textcolor{blue}{\footnotesize{}22.534} & {\footnotesize{}22.235} & \textbf{\footnotesize{}22.699}\tabularnewline
{\footnotesize{}SSIM} & {\footnotesize{}0.741} & {\footnotesize{}0.808} & {\footnotesize{}0.831} & \textbf{\footnotesize{}0.869} & \textcolor{blue}{\footnotesize{}0.865} & {\footnotesize{}0.683} & {\footnotesize{}0.642} & {\footnotesize{}0.758} & \textcolor{blue}{\footnotesize{}0.775} & {\footnotesize{}0.761} & \textbf{\footnotesize{}0.784}\tabularnewline
\midrule 
{\footnotesize{}LPIPS} & {\footnotesize{}0.489} & {\footnotesize{}0.345} & {\footnotesize{}0.323} & \textcolor{blue}{\footnotesize{}0.251} & \textbf{\footnotesize{}0.248} & {\footnotesize{}0.253} & {\footnotesize{}0.277} & {\footnotesize{}0.225} & \textcolor{blue}{\footnotesize{}0.220} & {\footnotesize{}0.225} & \textbf{\footnotesize{}0.202}\tabularnewline
{\footnotesize{}NIQE} & {\footnotesize{}11.338} & {\footnotesize{}7.850} & {\footnotesize{}7.770} & \textbf{\footnotesize{}6.311} & \textcolor{blue}{\footnotesize{}6.499} & {\footnotesize{}6.470} & \textbf{\footnotesize{}3.916} & {\footnotesize{}5.959} & {\footnotesize{}5.986} & {\footnotesize{}5.745} & \textcolor{blue}{\footnotesize{}5.055}\tabularnewline
\midrule 
{\footnotesize{}Static Loss} & {\footnotesize{}32.781} & {\footnotesize{}30.974} & \textcolor{blue}{\footnotesize{}31.032} & {\footnotesize{}30.824} & \textbf{\footnotesize{}31.085} & {\footnotesize{}25.415} & {\footnotesize{}24.303} & {\footnotesize{}26.323} & \textcolor{blue}{\footnotesize{}27.699} & {\footnotesize{}26.541} & \textbf{\footnotesize{}27.702}\tabularnewline
{\footnotesize{}Var. Dist.} & {\footnotesize{}21.762} & {\footnotesize{}23.383} & {\footnotesize{}23.730} & \textbf{\footnotesize{}24.342} & \textcolor{blue}{\footnotesize{}24.146} & {\footnotesize{}22.792} & {\footnotesize{}22.324} & {\footnotesize{}23.368} & \textcolor{blue}{\footnotesize{}23.478} & {\footnotesize{}23.423} & \textbf{\footnotesize{}23.587}\tabularnewline
{\footnotesize{}Warping err.} & {\footnotesize{}25.956} & \textbf{\footnotesize{}23.003} & \textcolor{blue}{\footnotesize{}22.624} & {\footnotesize{}21.856} & {\footnotesize{}22.207} & {\footnotesize{}19.873} & {\footnotesize{}19.156} & {\footnotesize{}20.194} & \textbf{\footnotesize{}20.915} & {\footnotesize{}20.308} & \textcolor{blue}{\footnotesize{}20.787}\tabularnewline
{\footnotesize{}tLPIPS} & {\footnotesize{}0.143} & {\footnotesize{}0.125} & {\footnotesize{}0.136} & \textbf{\footnotesize{}0.094} & \textcolor{blue}{\footnotesize{}0.098} & {\footnotesize{}0.504} & {\footnotesize{}0.866} & {\footnotesize{}0.658} & \textcolor{blue}{\footnotesize{}0.496} & {\footnotesize{}0.607} & \textbf{\footnotesize{}0.459}\tabularnewline
\bottomrule
\end{tabular}
\par\end{centering}
\vspace{1ex}

\caption{\label{tab:Experimental-validation}Experimental validation of our
proposed architecture for vid4 dataset. The table is separated into
pixel-error objective and perceptual objective methods. Best in bold
and runner-ups in blue (per category).}
\end{table*}
\begin{table}
\begin{centering}
\setlength{\tabcolsep}{4.2pt}
\begin{tabular}{lrrrrrr}
\toprule 
 & {\footnotesize{}bicubic} & {\footnotesize{}$\mathcal{L}_{E}$} & {\footnotesize{}ENet} & {\footnotesize{}SRGAN} & {\footnotesize{}$\mathcal{L}_{A}$} & {\footnotesize{}$\mathcal{L}_{C}$}\tabularnewline
\midrule 
{\footnotesize{}PSNR} & {\footnotesize{}27.131} & \textbf{\footnotesize{}29.750} & {\footnotesize{}25.131} & {\footnotesize{}23.918} & {\footnotesize{}26.411} & \textcolor{blue}{\footnotesize{}26.833}\tabularnewline
{\footnotesize{}SSIM} & {\footnotesize{}0.864} & \textbf{\footnotesize{}0.921} & {\footnotesize{}0.815} & {\footnotesize{}0.781} & {\footnotesize{}0.860} & \textcolor{blue}{\footnotesize{}0.869}\tabularnewline
\midrule 
{\footnotesize{}LPIPS} & {\footnotesize{}0.384} & {\footnotesize{}0.204} & {\footnotesize{}0.213} & {\footnotesize{}0.243} & \textcolor{blue}{\footnotesize{}0.154} & \textbf{\footnotesize{}0.146}\tabularnewline
{\footnotesize{}NIQE} & {\footnotesize{}3.895} & {\footnotesize{}3.814} & {\footnotesize{}4.970} & {\footnotesize{}4.141} & \textcolor{blue}{\footnotesize{}3.397} & \textbf{\footnotesize{}3.332}\tabularnewline
\midrule 
{\footnotesize{}Static Loss} & {\footnotesize{}32.046} & \textbf{\footnotesize{}31.309} & {\footnotesize{}27.300} & {\footnotesize{}26.341} & {\footnotesize{}27.941} & \textcolor{blue}{\footnotesize{}28.722}\tabularnewline
{\footnotesize{}Var. Dist.} & {\footnotesize{}24.590} & \textbf{\footnotesize{}26.877} & {\footnotesize{}25.448} & {\footnotesize{}24.692} & {\footnotesize{}25.977} & \textcolor{blue}{\footnotesize{}26.112}\tabularnewline
{\footnotesize{}Warping err.} & {\footnotesize{}21.825} & \textbf{\footnotesize{}20.296} & {\footnotesize{}19.243} & {\footnotesize{}19.059} & {\footnotesize{}19.319} & \textcolor{blue}{\footnotesize{}19.664}\tabularnewline
{\footnotesize{}tLPIPS} & {\footnotesize{}0.145} & \textbf{\footnotesize{}0.082} & \textcolor{blue}{\footnotesize{}0.383} & {\footnotesize{}0.622} & {\footnotesize{}0.525} & {\footnotesize{}0.417}\tabularnewline
\bottomrule
\end{tabular}
\par\end{centering}
\vspace{1ex}

\caption{Experimental evaluation of our proposed architecture for seq12 dataset.
Best in bold and runner-ups in blue.\label{tab:Experimental-evaluation-seq12}}

\end{table}

Our model falls in the category of recurrent neural networks, and
thus must be trained via Back-propagation Through Time (BPTT) \cite{Werbos1990},
which is a finite approximation of the infinite recurrent loop created
in the model. In practice, BPTT unfolds the network into several temporal
steps where each of those steps is a copy of the network sharing the
same parameters. The back-propagation algorithm is then used to obtain
gradients of the loss with respect to the parameters. We show an example
of unfolded recurrent generator and discriminator in Figure~\ref{fig:Unrolled-recurrent-network}.
We select $10$ temporal steps for our training approximation and
set the $n$ for our image alignment to~$5$. We choose a depth of
10 residual blocks for the image alignment and generator networks.

Our training set is composed by 4k videos downloaded from \textit{youtube.com}
and downscaled to $720\times1280$, from which we extract around 3M
$256\times256$ HR crops that serve as ground-truth images, and then
further downsample them by a factor of $s=4$ to obtain the LR input
of size $64\times64$. The training dataset thus is composed by around
300k sequences of 10 frames each (i.e.~around 300k data-points for
the recurrent network). We compile a testing set, larger than other
previous testing sets in the literature, also downloaded from \textit{youtube.com},
favoring sharp 4k content that is further downsampled to $720\times1280$
for GT and $180\times320$ for the LR input. In this dataset there
are 12 diverse sequences (e.g.~landscapes, natural wildlife, urban
scenes) ranging from very little to fast motion. Each sequence contains
100 to 150 frames (1281 frames in total).

We use a batch size of $8$ sequences, i.e.~each batch contains $8\times10=80$
training images. All models are pre-trained with $\mathcal{L}_{E}$
for about 2 epochs and then trained with the rest of the losses for
about 4 epochs more. The weights to the losses are:
\begin{equation}
\mathcal{L}_{c}=0.01\mathcal{L}_{E}+0.005\mathcal{L}_{A}+\mathcal{L}_{G}+0.1(\mathcal{L}_{T_{d}}+\mathcal{L}_{T_{s}}).\label{eq:L_c}
\end{equation}
Training was performed on Nvidia Tesla P100 and V100 GPUs, both of
which have 16 GB of memory.

\subsection{Evaluation}

\begin{figure*}
\setlength{\tabcolsep}{1.5pt}
\renewcommand{\arraystretch}{1.3}
\begin{centering}
\begin{tabular}{lccccc}
{\scriptsize{}\raisebox{15mm}{bic.}} & \includegraphics[width=0.18\textwidth]{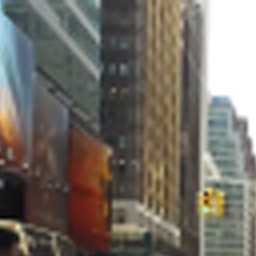} & \includegraphics[width=0.18\textwidth]{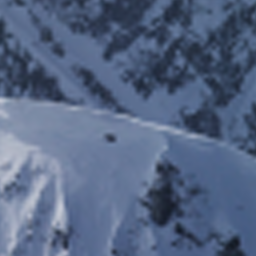} & \includegraphics[width=0.18\textwidth]{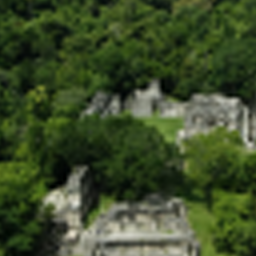} & \includegraphics[width=0.18\textwidth]{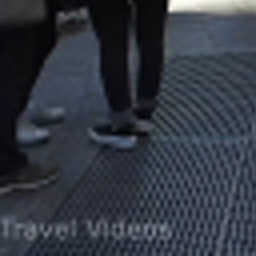} & \includegraphics[width=0.18\textwidth]{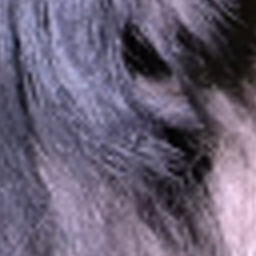}\tabularnewline
{\scriptsize{}\raisebox{15mm}{$\mathcal{L}_{E}$ (\ref{eq:L_e})}} & \includegraphics[width=0.18\textwidth]{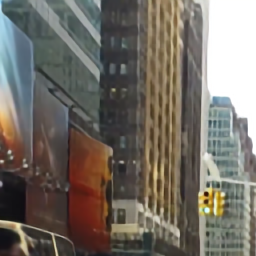} & \includegraphics[width=0.18\textwidth]{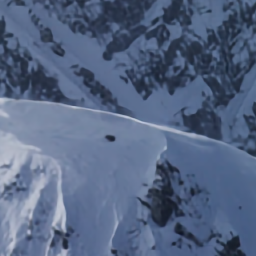} & \includegraphics[width=0.18\textwidth]{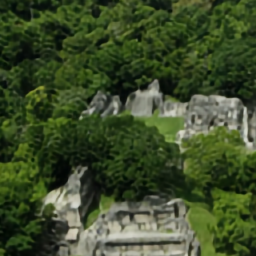} & \includegraphics[width=0.18\textwidth]{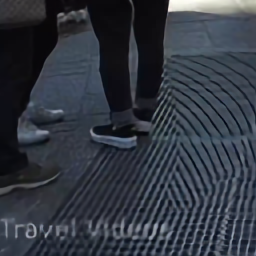} & \includegraphics[width=0.18\textwidth]{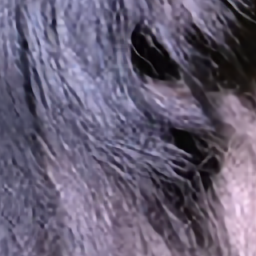}\tabularnewline
{\scriptsize{}\raisebox{15mm}{ENet \cite{SajSchHir17}}} & \includegraphics[width=0.18\textwidth]{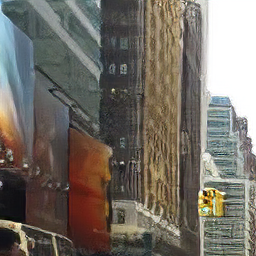} & \includegraphics[width=0.18\textwidth]{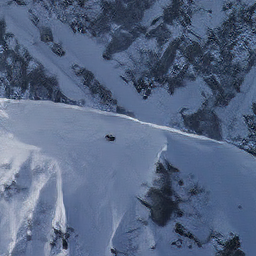} & \includegraphics[width=0.18\textwidth]{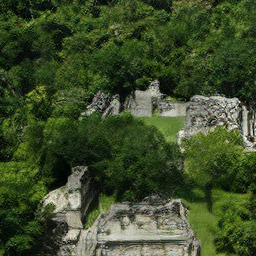} & \includegraphics[width=0.18\textwidth]{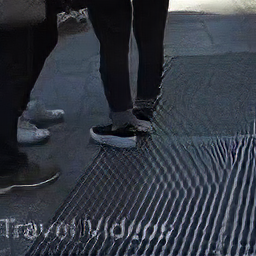} & \includegraphics[width=0.18\textwidth]{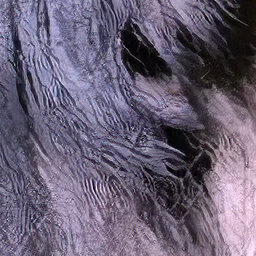}\tabularnewline
{\scriptsize{}\raisebox{15mm}{SRGAN\hspace{-7mm}}\raisebox{12mm}{\cite{Ledig2017}}} & \includegraphics[width=0.18\textwidth]{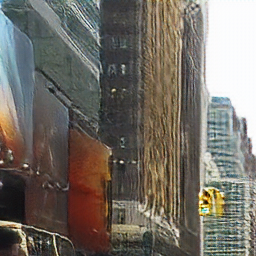} & \includegraphics[width=0.18\textwidth]{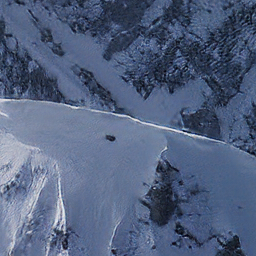} & \includegraphics[width=0.18\textwidth]{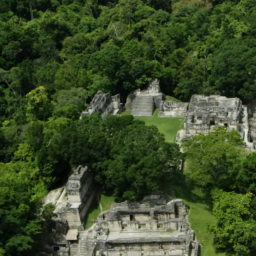} & \includegraphics[width=0.18\textwidth]{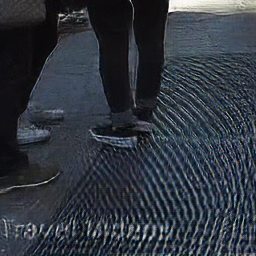} & \includegraphics[width=0.18\textwidth]{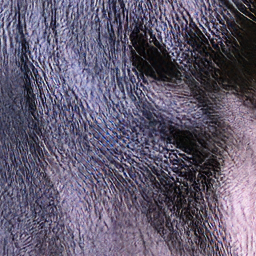}\tabularnewline
{\scriptsize{}\raisebox{15mm}{$\mathcal{L}_{C}$ (\ref{eq:L_c})}} & \includegraphics[width=0.18\textwidth]{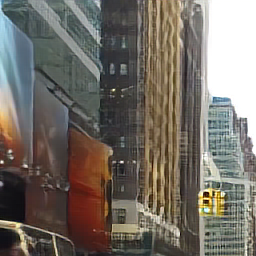} & \includegraphics[width=0.18\textwidth]{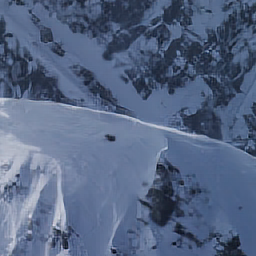} & \includegraphics[width=0.18\textwidth]{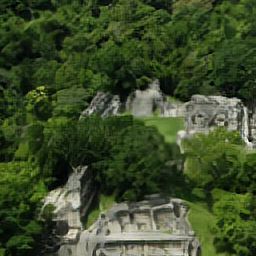} & \includegraphics[width=0.18\textwidth]{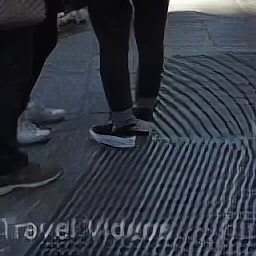} & \includegraphics[width=0.18\textwidth]{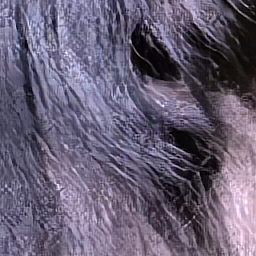}\tabularnewline
{\scriptsize{}\raisebox{15mm}{GT}} & \includegraphics[width=0.18\textwidth]{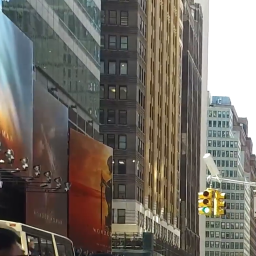} & \includegraphics[width=0.18\textwidth]{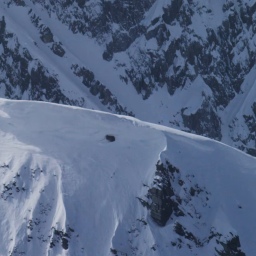} & \includegraphics[width=0.18\textwidth]{5_GT_crop34} & \includegraphics[width=0.18\textwidth]{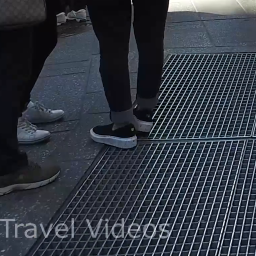} & \includegraphics[width=0.18\textwidth]{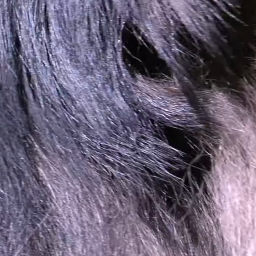}\tabularnewline
\end{tabular}
\par\end{centering}
\caption{Image close-ups for visual inspection and quantitative visual assessment.
The close-ups have been extracted from the following sequences (left
to right): \textit{newyork, mountain, tikal2, newyork, monkey}. \label{fig:Image-crops}}
\end{figure*}

\textbf{Models: }We performed exhaustive evaluation on intra-frame
quality and temporal consistency for vid4 and seq12 datasets. In Table~\ref{tab:Experimental-validation}
(right side) and Table~\ref{tab:Experimental-evaluation-seq12} we
compare perceptual image quality and temporal consistency metrics
against other generative SR methods, namely SRGAN \cite{Ledig2017}
(pretrained model obtained from \cite{srgan_tensorlayer}) and Enhancenet
\cite{SajSchHir17} (code and pre-trained network weights obtained
from the authors website). To the best of our knowledge, no perceptually
driven VSR methods have been published to date. In Table~\ref{tab:Experimental-validation}~(left
side) we also compare our methods with VSR methods based on MSE optimization:
Frame Recurrent Video Super-Resolution \cite{Sajjadi2018} (FRVSR),
Detail Revealing Deep VSR (DRDVSR) \cite{tao2017spmc}, and the Robust
VSR with learned temporal dynamics \cite{liu2017robust} (denoted
as $B_{1,2,3}+T$). For these methods, we compute the evaluation metrics
on the vid4 image results collected from the authors websites.

\textbf{Intra-frame quality}: Even though it is trivial for humans
to evaluate the perceived similarity between two images, the underlying
principles of human perception are still not well-understood. Traditional
metrics such as PSNR and Structural Self-Similarity (SSIM) still rely
on well-aligned, pixel-wise accurate estimates. In order to evaluate
image samples from models that deviate from the MSE minimization scheme
other metrics need to be considered.

Mittal et al.~\cite{Mittal_NIQE} introduced the no-reference Natural
Image Quality Evaluator (NIQE), which quantifies perceptual quality
by the deviation from natural image statistics in the spatial, wavelet
and DCT domains. Zhang et al.~\cite{Zhang2018} proposed recently
the Learned Perceptual Image Patch Similarity (LPIPS), which explore
the capabilities of deep architectures to capture perceptual features
that are meaningful for similarity assessment. In their exhaustive
evaluation they show how deep features of different architectures
outperform other previous metrics by substantial margins and correlate
very well with subjective human scores. They conclude that deep networks,
regardless of the specific architecture, capture important perceptual
features that are well-aligned with those of the human visual system.

We evaluate our testing sets with PSNR, SSIM, NIQE (we fit the NIQE
model to the GT image statistics of vid4 and seq12 separately) and
LPIPS using the AlexNet architecture with an additional linear calibration
layer as the authors propose in their manuscript. We show these scores
in Table~\ref{tab:Experimental-validation} and Table~\ref{tab:Experimental-evaluation-seq12},
and we show some image crops in Figure~\ref{fig:Image-crops} for
qualitative evaluation. Our method trained with $\mathcal{L}_{c}$
obtains the best LPIPS scores for both vid4 and seq12, and quantitative
inspection of the image crops suggest that even though SRGAN and Enhancenet
do generate fine textures, they also tend to deviate to a higher degree
from plausible texture patterns (e.g.~over sharpening in Figure~\ref{fig:Image-crops}
\textit{\small{}monkey}).

\textbf{Temporal Consistency}: Evaluating the temporal consistency
over adjacent frames in a sequence where the ground-truth optical
flow is not known is an open problem. It is common in the literature
to compute the warping error across consecutive frames with a flow
estimator \cite{Gupta2017,Huang2017,Lai-ECCV-2018}. We compute the
warping error PSNR with flow estimates from PWC-Net \cite{Sun2018PWC-Net}.
Please refer to the supplementary material for further discussion
on this metric.

We include as well the tLPIPS as proposed by \cite{Mengyu_1811.09393},
which computes the LPIPS distance for consecutive estimated frames
and references it to that of the ground-truth images $\textrm{tLPIPS}=\left\Vert \Lambda(\tilde{X}_{t-1},\tilde{X}_{t})-\Lambda(X_{t-1},X_{t})\right\Vert _{1},$
where $\Lambda(X_{t-1},X)$ computes the LPIPS score between image
$X_{t-1}$ and $X_{t}$. Additionally, we evaluate temporal consistency
with $\mathcal{L}_{T_{d}}$ (i.e.~static loss) and $\mathcal{L}_{T_{s}}$
(i.e.~variance distance), both of them related to reliable ground-truth
data and with a simple interpretation (motionless flickering and differences
in pixel statistics across time). For a clearer comparison, we present
these two temporal metrics in logarithmic scale. We show the results
in Table~\ref{tab:Experimental-validation} and \ref{tab:Experimental-evaluation-seq12}.
As expected, there is a gap between methods optimized with pixel distances
(e.g.~$\mathcal{L}_{E}$, FRVSR) and generative algorithms. Within
the later, all the configurations of our model perform well in all
temporal metrics, even when we do not minimize any of the proposed
temporal losses directly (i.e.~$\mathcal{L}_{A}$), which supports
the effectiveness of the video discriminator and the recurrent generator.
$\mathcal{L}_{c}$ is the best performer among the perceptual methods.
In contrast, models that are not aware of the temporal dimensions
(such as Enhacenet or SRGAN) obtain worse scores, being SRGAN the
worst performer (which is in line with what the quantitative evaluation
of video sequences suggest).

\textbf{Ablation Study}: We show in Table~\ref{tab:Experimental-validation}
an ablation study of our proposed architecture trained with different
loss functions: $\mathcal{L}_{E}$, $\mathcal{L}_{A}$, $\mathcal{L}_{T_{d}}$,
$\mathcal{L}_{T_{s}}$ and $\mathcal{L}_{C}$. Firstly, we would like
to remark that $\mathcal{L}_{A}$ includes a video discriminator,
and thus its temporal consistency is improved when compared to SRGAN
or Enhancenet. If we compare $\mathcal{L}_{E}$ and $\mathcal{L}_{A}$
we observe how perceptual quality metrics improve, however all temporal
consistency metrics degrade (i.e.~temporal consistency in perceptually
driven methods is challenging). When comparing separately $\mathcal{L}_{T_{d}}$
and $\mathcal{L}_{T_{s}}$ to $\mathcal{L}_{A}$, we observe that
both loss terms improve temporal consistency and quality metrics,
which suggest that temporal consistency helps obtaining better intra-frame
quality as well. The Static Loss $\mathcal{L}_{T_{d}}$ has a higher
impact on the temporal consistency metrics than $\mathcal{L}_{T_{s}}$.
Finally, when optimizing our proposed loss function $\mathcal{L}_{C}$
we further improve temporal consistency and quality over both $\mathcal{L}_{T_{d}}$
and $\mathcal{L}_{T_{s}}$.

\textbf{}

\section{Conclusions}

We present a novel generative adversarial model for video upscaling.
Differently from previous approaches to video super-resolution based
on MSE minimization, we use an adversarial loss function in order
to recover videos with photorealistic textures. To the best of our
knowledge, this is the first work that applies perceptual loss functions
to the task of video super-resolution.

In order to tackle the problem of lacking temporal consistency, we
propose three contributions: (1) A recurrent generative adversarial
model with a video discriminator, (2)~a multi-image warping that
improves image alignment between adjacent frames, and (3) two novel
loss terms that reinforce temporal coherency for consecutive frames.
We conducted exhaustive quantitative evaluation both for intra-frame
quality and temporal consistency on vid4 and seq12 (1281 frames) datasets.
Our method obtains state-of-the-art results in terms of LPIPS and
NIQE scores, and it improves temporal consistency when compared to
other generative models such as SRGAN or Enhacenet.

\bibliographystyle{ieee}
\bibliography{egbib}

\begin{thebibliography}{10}\itemsep=-1pt

\bibitem{ntire}
E.~Agustsson and R.~Timofte.
\newblock {NTIRE} 2017 challenge on single image super-resolution: Dataset and
  study.
\newblock In {\em {CVPR} workshop}, 2017.

\bibitem{Caballero2017}
J.~Caballero, C.~Ledig, A.~Aitken, A.~Acosta, J.~Totz, Z.~Wang, and W.~Shi.
\newblock Real-time video super-resolution with spatio-temporal networks and
  motion compensation.
\newblock In {\em CVPR}, 2017.

\bibitem{udnet}
C.~Chen, X.~Tian, F.~Wu, and Z.~Xiong.
\newblock {UDNet}: Up-down network for compact and efficient feature
  representation in image super-resolution.
\newblock In {\em ICCV}, 2017.

\bibitem{Mengyu_1811.09393}
M.~Chu, Y.~Xie, L.~Leal-Taix\'e, and N.~Thuerey.
\newblock Temporally coherent gans for video super-resolution (tecogan), 2018.

\bibitem{srcnn}
C.~Dong, C.~C. Loy, K.~He, and X.~Tang.
\newblock Learning a deep convolutional network for image super-resolution.
\newblock In {\em ECCV}, 2014.

\bibitem{fastdong}
C.~Dong, C.~C. Loy, and X.~Tang.
\newblock Accelerating the super-resolution convolutional neural network.
\newblock In {\em ECCV}, 2016.

\bibitem{srgan_tensorlayer}
H.~Dong, A.~Supratak, L.~Mai, F.~Liu, A.~Oehmichen, S.~Yu, and Y.~Guo.
\newblock {SRGAN} implementation, tensorlayer.
\newblock \url{https://github.com/tensorlayer/srgan/releases/tag/1.2.0}.

\bibitem{brox}
A.~Dosovitskiy and T.~Brox.
\newblock Generating images with perceptual similarity metrics based on deep
  networks.
\newblock In {\em NIPS}, 2016.

\bibitem{Dosovitskiy:2015:FLO:2919332.2919957}
A.~Dosovitskiy, P.~Fischery, E.~Ilg, P.~Hausser, C.~Hazirbas, V.~Golkov,
  P.~v.~d. Smagt, D.~Cremers, and T.~Brox.
\newblock Flownet: Learning optical flow with convolutional networks.
\newblock In {\em ICCV}, 2015.

\bibitem{Gatys2016}
L.~A. Gatys, A.~S. Ecker, and M.~Bethge.
\newblock Image style transfer using convolutional neural networks.
\newblock In {\em CVPR}, 2016.

\bibitem{Goodfellow2014}
I.~Goodfellow, J.~Pouget-Abadie, M.~Mirza, B.~Xu, D.~Warde-Farley, S.~Ozair,
  A.~Courville, and Y.~Bengio.
\newblock Generative adversarial nets.
\newblock In {\em NIPS}, 2014.

\bibitem{Gupta2017}
A.~Gupta, J.~Johnson, A.~Alahi, and L.~Fei-Fei.
\newblock Characterizing and improving stability in neural style transfer.
\newblock In {\em ICCV}, 2017.

\bibitem{residualnets}
K.~He, X.~Zhang, S.~Ren, and J.~Sun.
\newblock Deep residual learning for image recognition.
\newblock In {\em CVPR}, 2016.

\bibitem{Huang2017}
H.~Huang, H.~Wang, W.~Luo, L.~Ma, W.~Jiang, X.~Zhu, Z.~Li, and W.~Liu.
\newblock Real-time neural style transfer for videos.
\newblock In {\em CVPR}, 2017.

\bibitem{bidirectional}
Y.~Huang, W.~Wang, and L.~Wang.
\newblock Bidirectional recurrent convolutional networks for multi-frame
  super-resolution.
\newblock In {\em NIPS}, 2015.

\bibitem{johnson}
J.~Johnson, A.~Alahi, and L.~Fei-Fei.
\newblock Perceptual losses for real-time style transfer and super-resolution.
\newblock In {\em ECCV}, 2016.

\bibitem{vsrnet}
A.~Kappeler, S.~Yoo, Q.~Dai, and A.~K. Katsaggelos.
\newblock Video super-resolution with convolutional neural networks.
\newblock In {\em IEEE Transactions on Computational Imaging}, 2016.

\bibitem{vdsr}
J.~Kim, J.~Kwon~Lee, and K.~Mu~Lee.
\newblock Accurate image super-resolution using very deep convolutional
  networks.
\newblock In {\em CVPR}, 2016.

\bibitem{drcn}
J.~Kim, J.~Kwon~Lee, and K.~Mu~Lee.
\newblock Deeply-recursive convolutional network for image super-resolution.
\newblock In {\em CVPR}, 2016.

\bibitem{Kim2016}
J.~Kim, J.~K. Lee, and K.~M. Lee.
\newblock Accurate image super-resolution using very deep convolutional
  networks.
\newblock In {\em CVPR}, 2016.

\bibitem{lapsr}
W.-S. Lai, J.-B. Huang, N.~Ahuja, and M.-H. Yang.
\newblock Deep laplacian pyramid networks for fast and accurate
  super-resolution.
\newblock In {\em CVPR}, 2017.

\bibitem{Lai-ECCV-2018}
W.-S. Lai, J.-B. Huang, O.~Wang, E.~Shechtman, E.~Yumer, and M.-H. Yang.
\newblock Learning blind video temporal consistency.
\newblock In {\em European Conference on Computer Vision}, 2018.

\bibitem{Ledig2017}
C.~Ledig, L.~Theis, F.~Husz{\'a}r, J.~Caballero, A.~Cunningham, A.~Acosta,
  A.~Aitken, A.~Tejani, J.~Totz, Z.~Wang, and W.~Shi.
\newblock Photo-realistic single image super-resolution using a generative
  adversarial network.
\newblock In {\em CVPR}, 2017.

\bibitem{Li2017}
D.~Li, Y.~Liu, and Z.~Wang.
\newblock Video super-resolution using motion compensation and residual
  bidirectional recurrent convolutional network.
\newblock In {\em ICIP}, 2017.

\bibitem{Lim2017}
B.~Lim, S.~Son, H.~Kim, S.~Nah, and K.~M. Lee.
\newblock Enhanced deep residual networks for single image super-resolution.
\newblock In {\em {CVPR} workshop}, 2017.

\bibitem{bayesianadaptive}
C.~Liu and D.~Sun.
\newblock A bayesian approach to adaptive video super resolution.
\newblock In {\em CVPR}, 2011.

\bibitem{robustdyn}
D.~Liu, Z.~Wang, Y.~Fan, X.~Liu, Z.~Wang, S.~Chang, and T.~Huang.
\newblock Robust video super-resolution with learned temporal dynamics.
\newblock In {\em CVPR}, 2017.

\bibitem{liu2017robust}
D.~Liu, Z.~Wang, Y.~Fan, X.~Liu, Z.~Wang, S.~Chang, and T.~Huang.
\newblock Robust video super-resolution with learned temporal dynamics.
\newblock In {\em Proceedings of the IEEE Conference on Computer Vision and
  Pattern Recognition}, pages 2507--2515, 2017.

\bibitem{endtoendmotion}
O.~Makansi, E.~Ilg, and T.~Brox.
\newblock End-to-end learning of video super-resolution with motion
  compensation.
\newblock In {\em GCPR}, 2017.

\bibitem{peyman}
P.~Milanfar.
\newblock {\em Super-resolution Imaging}.
\newblock CRC press, 2010.

\bibitem{Mittal_NIQE}
A.~{Mittal}, R.~{Soundararajan}, and A.~C. {Bovik}.
\newblock Making a completely blind image quality analyzer.
\newblock {\em IEEE Signal Processing Letters}, 20(3), 2013.

\bibitem{survey14}
K.~Nasrollahi and T.~B. Moeslund.
\newblock Super-resolution: A comprehensive survey.
\newblock {\em Machine Vision and Applications}, 2014.

\bibitem{Radford2016}
A.~Radford, L.~Metz, and S.~Chintala.
\newblock Unsupervised representation learning with deep convolutional
  generative adversarial networks.
\newblock In {\em ICLR}, 2016.

\bibitem{spynet}
A.~Ranjan and M.~Black.
\newblock Optical flow estimation using a spatial pyramid network.
\newblock In {\em Proceedings IEEE Conference on Computer Vision and Pattern
  Recognition (CVPR) 2017}, Piscataway, NJ, USA, July 2017. IEEE.

\bibitem{Ruder2016}
M.~Ruder, A.~Dosovitskiy, and T.~Brox.
\newblock Artistic style transfer for videos.
\newblock {\em GCPR}, 2016.

\bibitem{SajSchHir17}
M.~S.~M. {Sajjadi}, B.~Sch{\"o}lkopf, and M.~Hirsch.
\newblock {EnhanceNet}: Single image super-resolution through automated texture
  synthesis.
\newblock In {\em ICCV}, 2017.

\bibitem{Sajjadi2018}
M.~S.~M. {Sajjadi}, R.~{Vemulapalli}, and M.~{Brown}.
\newblock {Frame-Recurrent Video Super-Resolution}.
\newblock In {\em CVPR}, 2018.

\bibitem{Shi2016}
W.~Shi, J.~Caballero, F.~Husz\'ar, J.~Totz, A.~P. Aitken, R.~Bishop,
  D.~Rueckert, and Z.~Wang.
\newblock Real-time single image and video super-resolution using an efficient
  sub-pixel convolutional neural network.
\newblock In {\em CVPR}, 2016.

\bibitem{Simonyan15}
K.~Simonyan and A.~Zisserman.
\newblock Very deep convolutional networks for large-scale image recognition.
\newblock In {\em International Conference on Learning Representations}, 2015.

\bibitem{Sun2018PWC-Net}
D.~Sun, X.~Yang, M.-Y. Liu, and J.~Kautz.
\newblock {PWC-Net}: {CNNs} for optical flow using pyramid, warping, and cost
  volume.
\newblock In {\em CVPR}, 2018.

\bibitem{deepresnetssr}
Y.~Tai, J.~Yang, and X.~Liu.
\newblock Image super-resolution via deep recursive residual network.
\newblock In {\em CVPR}, 2017.

\bibitem{detailrevealing}
X.~Tao, H.~Gao, R.~Liao, J.~Wang, and J.~Jia.
\newblock Detail-revealing deep video super-resolution.
\newblock In {\em ICCV}, 2017.

\bibitem{tao2017spmc}
X.~Tao, H.~Gao, R.~Liao, J.~Wang, and J.~Jia.
\newblock Detail-revealing deep video super-resolution.
\newblock In {\em The IEEE International Conference on Computer Vision (ICCV)},
  Oct 2017.

\bibitem{Werbos1990}
P.~J. Werbos.
\newblock Backpropagation through time: what it does and how to do it.
\newblock {\em Proceedings of the IEEE}, pages 1550--1560, 1990.

\bibitem{Xiang2017}
S.~Xiang and H.~Li.
\newblock On the effects of batch and weight normalization in generative
  adversarial networks.
\newblock {\em arXiv:1704.03971v4}, 2017.

\bibitem{Zhang2018}
R.~Zhang, P.~Isola, A.~A. Efros, E.~Shechtman, and O.~Wang.
\newblock The unreasonable effectiveness of deep features as a perceptual
  metric.
\newblock {\em arXiv:1801.03924}, 2018.

\end{thebibliography}

\end{document}